\pdfoutput=1
\documentclass[11pt]{article}

\usepackage{EMNLP2023}

\usepackage{amsfonts}
\usepackage{times}
\usepackage{latexsym}
\usepackage{tabularx}
\usepackage{ulem}
\usepackage{graphicx} 
\usepackage{amsmath}
\usepackage[T1]{fontenc}
\usepackage[utf8]{inputenc}

\usepackage{microtype}

\usepackage{inconsolata}

\newcommand{\sysname}{\textsc{SCALE}}

\title{Fast and Accurate Factual Inconsistency Detection Over Long Documents}

\author{Barrett Martin Lattimer \\
ASAPP \\
\texttt{blattimer@asapp.com} \\\And
{\bf Patrick Chen} \\
ASAPP \\
\texttt{pchen@asapp.com} \\\And
Xinyuan Zhang\thanks{\,\,Work done while at ASAPP.} \\
Abstractive Health \\
\texttt{dylanz0426@gmail.com} \\\AND
Yi Yang \\
ASAPP \\
\texttt{yyang@asapp.com}\\}

\begin{document}
\maketitle
\begin{abstract}
%Generative AI models demonstrate remarkable potential, but hallucinations across various tasks pose a significant challenge, especially when dealing with longer inputs which current factual inconsistency metrics struggle to handle effectively. We introduce \sysname~(Source Chunking Approach for Large-scale consistency Evaluation), a task-agnostic factual inconsistency detection model that uses a novel chunking approach to achieve state-of-the-art hallucination detection performance over various tasks and long inputs. \sysname~is a Natural Language Inference (NLI) based model that uses large text chunks to condition over long texts.  Furthermore, we present a novel approach based on the chunking mechanism to explain the decisions of \sysname using relevant source sentence retrieval. 
%Our evaluations shown that \sysname~outperforms existing methods on both standard benchmarks for factual inconsistency detection and a new dataset that features long-form dialogues constructed by us. In addition, \sysname~is also significantly better than competitive systems in efficiency and model explanation evaluations.
%shown to have faster processing speeds and improved score calibration across multiple NLG tasks. 
%To provide a robust testing ground for \sysname~and future metrics, we present ScreenEval, a new long-form dialogue-based dataset featuring unique challenges in factual inconsistency detection.
%We will release our code and data upon the acceptance of the paper.

Generative AI models exhibit remarkable potential; however, hallucinations across various tasks present a significant challenge, particularly for longer inputs that current approaches struggle to address effectively. We introduce \sysname~(\textbf{S}ource \textbf{C}hunking \textbf{A}pproach for \textbf{L}arge-scale inconsistency \textbf{E}valuation), a task-agnostic model for detecting factual inconsistencies using a novel chunking strategy. Specifically, \sysname~is a Natural language inference (NLI) based model that uses large text chunks to condition over long texts. This approach achieves state-of-the-art performance in factual inconsistency detection for diverse tasks and long inputs. Additionally, we leverage the chunking mechanism and employ a novel algorithm to explain \sysname's decisions through relevant source sentence retrieval. Our evaluations reveal that \sysname~outperforms existing methods on both standard benchmarks and a new long-form dialogue dataset ScreenEval we constructed. Moreover, \sysname~surpasses competitive systems in efficiency and model explanation evaluations. We have released our code and data publicly to GitHub\footnote{\url{https://github.com/asappresearch/scale-score}}.

% Generative AI models are powerful, however, one of the biggest challenges is hallucination across a multitude of tasks.
% Additionally, many of these NLG tasks take long inputs however modern factual inconsistency metrics are not able to handle arbitrarily long inputs well. 
% We present SCALE (Source Chunking Approach for Large-scale consistency Evaluation), a task agnostic model based factual inconsistency metric that uses a novel chunking approach to achieve state of the art performance in hallucination detection efficiently over long inputs. 
% SCALE is a Natural Language Inference (NLI) based model that uses large chunks of text to condition on long texts. 
% We show that SCALE achieves superior results over other existing metrics across 5 different NLG tasks. In addition, SCALE is able to perform state of the art relevant sentence retrieval at a fraction of the speed needed by other models and has superior score calibration across many NLG tasks.
% We introduce a novel long form dialogue based dataset ScreenEval that presents unique challenges in factual inconsistency detection to existing and future metrics.
% We release our code and dataset publicly.
\end{abstract}

%%% You are an academic paper writer specialized in natural language processing.
%%% Please polish the following paragraph in the introduction section. Please make it concise.

\section{Introduction}
% LLMs are powerful, however, one of the biggest challenges in deploying LLMs in real time applications is hallucinations.
% Previous efforts have drawbacks: slow (not sutible for real time apps, not accurate, or little calibration)
% To address these drawbacks we come up with SCALE
% In our efforts, there is no good dataset
% Experiments.

% YI: outline
% 1. Generative AI and Large Language Models, such as OpenAI ChatGPT/GPT-4, Anthropic Claude, and Google BARD, are powerful tools and have numerous potentials; One of the most popular and impressive capabilities is that they can rapidly read and understand a document up to 100k tokens and summarize or answer any questions about it; Unfortunately, hallucinations, the How to quickly validate the 

% 2. Previous efforts focus on short documents, 

Large Language Models (LLMs) have shown immense promise in various applications, but deploying them in real-time presents certain challenges such as hallucinations \cite{cao2018faithful, falke2019ranking, kryscinski2019evaluating, fabbri2021summeval, honovich2022true}. Hallucinations, or factual inconsistencies generated by a model relative to a source document, can mislead the user and undermine trust in LLMs. Thus, detecting factual inconsistency in LLM generations is crucial for the future of LLMs, especially with the growing popularity of platforms like ChatGPT.

% Yi's draft
% Previous efforts in hallucination detection have been focused on short documents~\cite{} in the offline setting. They are slow in processing long documents and poorly calibrated. This poses significant challenges in adopting them in the real world online setting, as introducing hallucination detection could lead to an order of magnitude increase in latency and the lack of well calibrated scores makes it tricky to balance the risk of including hallucination and excluding salient information in the model output. Considering the exponentially increasing context sizes (maximum allowed tokens of an input) of recent LLMs.\footnote{For example, OpenAI GPT-4 and Anthropic Claude have supported context sizes up to 32k and 100k respectively.}, the need of efficient and effective approach for hallucination detection on long documents becomes more and more pressing. 

% GPT-4 polished
Prior research on inconsistency detection has predominantly dealt with short documents in offline settings~\cite{laban2022summac, schuster2022stretching, utama2022falsesum} and relied heavily on sentence-level text matching techniques. Consequently, these methods exhibit slow performance in processing longer documents and suffer from poor calibration. Such characteristics pose substantial challenges in implementing them in real-world online environments, where incorporating inconsistency detection could potentially result in a substantial increase in latency. Additionally, the absence of well-calibrated scores complicates the balancing act between mitigating the risk of incorporating hallucinations and excluding pertinent information from the model output. Given the exponential growth in context sizes (maximum allowed tokens of an input) of contemporary large language models (LLMs),\footnote{For instance, OpenAI GPT-4 and Anthropic Claude support context sizes up to 32k and 100k tokens, respectively.} there is an increasing urgency to develop efficient and effective approaches for inconsistency detection in lengthy documents.

In addressing the challenges, we introduce \sysname~(\textbf{S}ource \textbf{C}hunking \textbf{A}pproach for \textbf{L}arge-scale inconsistency \textbf{E}valuation), a method designed for efficient detection of factual inconsistencies in generated sentences by identifying related source text snippets. \sysname~consists of two crucial components. First, it builds on a Natural language inference (NLI) based method, integrating a novel chunking mechanism for rapid and accurate online performance in diverse natural language generation (NLG) tasks. Second, model explanation is essential for real-time deployment of inconsistency detection systems, facilitating swift human inspection to determine model configurations. We show that our chunking mechanism improves calibration scores and enables the use of a binary search tree algorithm for rapidly locating relevant source text snippets for a target sentence, ultimately enhancing the explanation of model behaviors.

%Existing benchmark datasets for factual inconsistency detection contains only short documents. To evaluate SCALE on a realistic dataset with lengthy documents, we build ScreenEval, a new dataset that evaluates the factual inconsistency of summary sentences generated by humans, Longformer, and GPT-4 against real long-form dialogues. ScreenEvals contains 52 dialogues with an average of 5-6k tokens per dialogue. It is a challenging dataset due to long coreference resolution and significant noise between utterances. To the best of our knowledge, it is the largest dialogue-based factual inconsistency detection dataset. 

Current benchmark datasets for factual inconsistency detection predominantly feature short documents. In order to evaluate SCALE using a more realistic dataset with long documents, we introduce ScreenEval — a novel dataset designed to assess the factual inconsistency of summary sentences generated by humans, Longformer, and GPT-4 in comparison to actual long-form dialogues. ScreenEval encompasses 52 dialogues, averaging over 6,000 tokens per dialogue. The use of dialogue in this dataset poses a considerable unique challenges such as long-distance coreference resolution and significant noise between utterances. To the best of our knowledge, ScreenEval is the longest dialogue based dataset for factual inconsistency detection presently available.

In our experiments, we first show that \sysname~outperforms and is better calibrated than baseline methods across various NLG tasks on the standard factual inconsistency detection benchmark TRUE~\cite{honovich2022true}. We then assess accuracy, speed, and model explanation (via relevant text retrieval evaluation) on the new ScreenEval dataset for long document factual inconsistency detection. Our findings indicate that \sysname~surpasses strong competitors in the majority of tests. The key contributions of this paper are:
\begin{itemize}
    \item We introduce \sysname, a reference-free, NLI based factual inconsistency detection method with a novel chunking strategy for versatility across domains and extended documents. 
    \item We show \sysname's broad applicability in NLG domains by attaining state-of-the-art performance on the TRUE benchmark.
    \item We build ScreenEval, a novel dataset designed for factual inconsistency detection in long dialogues, and then demonstrate \sysname's superiority in accuracy, efficiency, and model explanation evaluations on the dataset.
\end{itemize}

\begin{figure*}[th!]
    \centering
    \includegraphics[width=\textwidth]{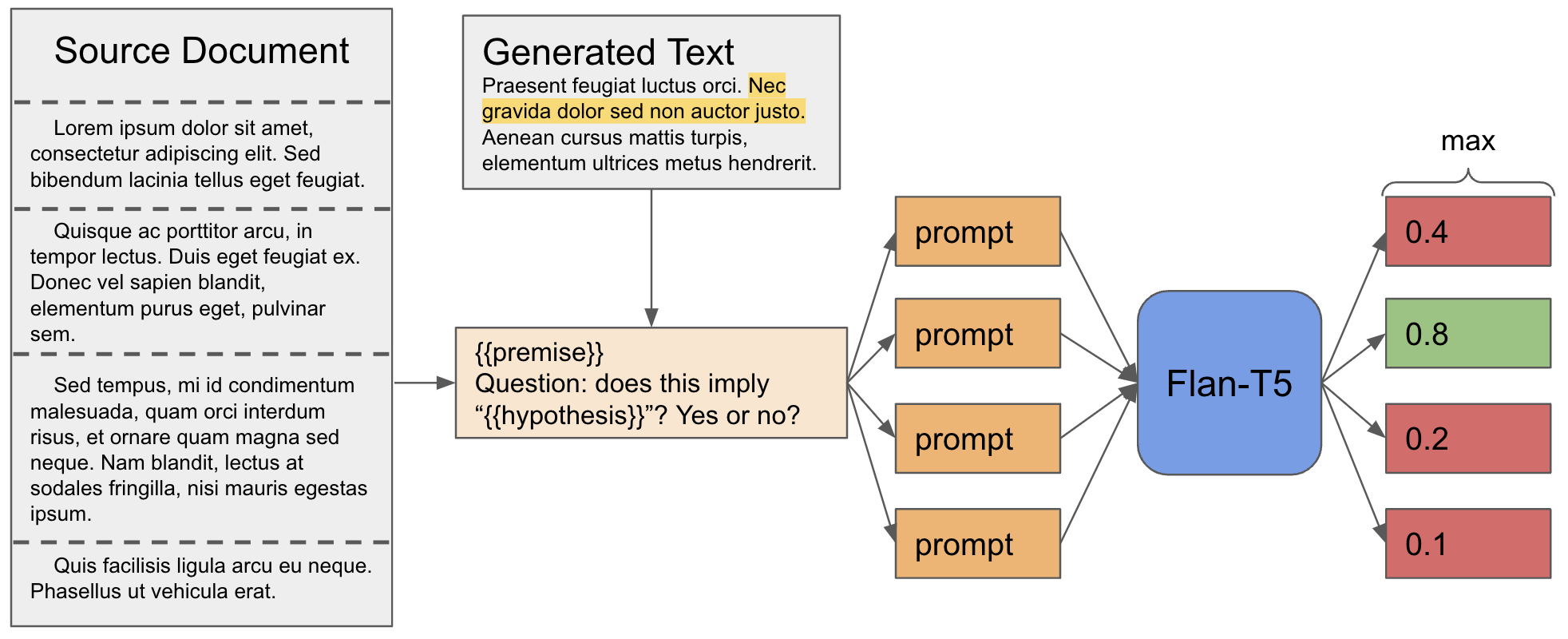}
    \caption{Chunking mechanism for \sysname~to produce a score given a source document and generated text. The source document is broken into chunks (represented by dashed lines) and each chunk is fed into to a prompt as the premise. The highlighted generated text is fed into all prompts as the hypothesis. Each prompt is then run through Flan-T5 and the resulting logits are used to compute the entailment score.}
    \label{fig:chunking}
    \vspace{-1em}
\end{figure*}
\vspace{-.8em}
\section{Related Work}

\paragraph{Factual Inconsistency Detection}

There are two main directions in factual inconsistency detection: Natural language inference (NLI) based and question answering (QA) based methods. In NLI based methods, pretrained NLI models can be utilized to determine whether a given "premise" factually entails a "hypothesis." Although initial attempts encountered challenges~\cite{khot2018scitail}, recent advancements have shown that NLI models can effectively assess the factual consistency of generated text (hypothesis) with respect to a source (premise) \cite{utama2022falsesum}. This progress can largely be attributed to addressing the granularity problem, which arises from the abundance of current NLI datasets predominantly comprised of short, single-sentence premises and hypotheses \cite{williams2017broad, nie2019adversarial, thorne2018fever, schuster2021get}.

% approaches have dominated the literature on the task of factual inconsistency detection. 

% Numerous benchmarks and methodologies have been proposed, which involve decomposing the source document and generated text into sentences and conducting pairwise comparisons between them \cite{huang2023swing, schuster2022stretching, laban2022summac, yin2021docnli}. While this strategy yields improved performance compared to evaluating the entire generated text simultaneously, it compromises the context from the source document. Prior research has determined that no optimal approach exists for decomposing source documents and hypotheses in the NLI task \cite{glover2022revisiting}; however, these studies only compared sentence decomposition against non-decomposition. Other investigations concluded that employing longer premises rather than single sentences is detrimental \cite{yin2021docnli}. In this paper, we further explore this area by examining an additional level of granularity: chunks.

\sysname~is an NLI based method and our findings indicate that utilizing larger premise chunks enhances efficiency and outperforms sentence decomposition. Although SeNtLI \cite{schuster2022stretching} extended NLI based methods to longer documents, it adhered to the sentence decomposition assumption and focused solely on summarization tasks. SummaC \cite{laban2022summac} investigated various aggregation techniques for NLI scores obtained from sentence decomposition to generate overall summary scores. Meanwhile, SWING \cite{huang2023swing} developed a loss function to train models for improved NLI performance, yielding mixed outcomes.

In QA based methods, a question is first generated based on a summary sentence, and a QA system is used to give an answer. A summary is considered factually consistent if the generated answer significantly overlaps with the original summary \cite{durmus2020feqa}. Prior research focused on using different question generation strategies \cite{scialom2019answers} or overlap measures \cite{deutsch2021understanding}. In the experiments, we consider the most competitive QuestEval \cite{scialom2021questeval} and QAFactEval \cite{fabbri2021qafacteval}.

\paragraph{Detecting Factual Inconsistencies in Long Documents}
%Current NLI datasets are almost all limited to short premises and hypothesis, limiting the use of NLI on longer documents. 
%Additionally, many NLI datasets are variants of the CNN/DailyMail news summary dataset which causes a lack of diversity in current NLI datasets and benchmarks \cite{kryscinski2019evaluating, fabbri2021summeval}. 

%Evaluating consistency in dialogue poses unique challenges which were approached early on by \cite{wang2022analyzing}, who focused on short dialogues and summaries from the SamSum corpus \cite{gliwa2019samsum}. 
% However, long-form dialogue evaluation is important due to the complex coreference resolution and noise that results in a more difficult task \cite{chen2021summscreen}. 

% Although some previous works have explored factual inconsistency performance on long source documents, their scope has been limited.
%For instance, ContractNLI \cite{koreeda2021contractnli} only focuses on legal documents which do not pose the same challenges as dialogue. 
%Similarly, LongEval \cite{krishna2023longeval} explores consistency detection over long documents however it focuses on the human evaluation strategies for scoring such documents and also does not evaluate dialogue. 
%To the best of our knowledge this paper provides the first factual inconsistency long form dialogue evaluation dataset ScreenEval filling a considerable gap in the literature. 

Prior work on factual inconsistency performance in long source documents has been limited in scope. For example, ContractNLI \cite{koreeda2021contractnli} concentrates on legal documents, which differ significantly from dialogues in terms of challenges. Likewise, LongEval \cite{krishna2023longeval} emphasizes human evaluation strategies for scoring, without considering dialogues. To our knowledge, this paper presents the first dataset for evaluating factual inconsistency in long-form dialogues, ScreenEval, thus addressing a substantial gap in the literature.

\vspace{-.1em}
\section{\sysname}
In this section we elaborate on our approach taken for our inconsistency detection model \sysname.
% In this section, we explore two ideas to speed up and improve the accuracy of factual inconsistency detection. 
Firstly, we formally define the use of chunks and NLI in \sysname, aiming at improving the accuracy and efficiency of the system. Secondly, we propose to explain the model output by retrieving the relevant source sentence for a target, and show how the relevant sentence retrieval can be improved through the use of chunks. 
%This section seeks to rigorously define each of these approaches. 

% \subsection{Consistency Detection}
% We define a generated text to be consistent with respect to it’s source document if the source document factually implies the entirety of the text.
% We take a close world approach to consistency, only relying on the facts in the source document to decide if generated text is consistent or not. 
% Formally defined, given $S$ as the set of facts in the source document and $G$ as the set of facts in the generated text, a generated text is considered consistent with respect to the source document if $G \subseteq S$. \pat{This paragraph is a bit unneccesary. Should define more stuff to make it valid. In particular, is $G$ or $S$ defined here used anywhere? I deliberately add something in below but did you use it anywhere else?}
% \xz{We can remove the subsection and define the problem above (under section 3)}

\subsection{Chunking Mechanism for NLI based Model}
Our approach uses NLI \cite{dagan2006pascal} as a building block for factual inconsistency detection. An NLI model $M$ provides the relationship between a premise $p$ and a hypothesis $h$, $M(p,h)$ with probabilities of three labels: entailed, neutral, and contradictory. For example, given an NLI model $M(p,h)$, source document $D$ with a set of facts $F_D$, and a generated text $G$ with a set of facts $F_G$, if $F_G\subseteq F_D$ we would expect $M(D,G)$ to produce high entailment probability. 

We define factual consistency between a generated text and source document as $F_G\subseteq F_D$. 
Canonical NLI models cannot be properly used for factual inconsistency detection because both $p$ and $h$ are commonly single sentences in NLI models, however in the factual inconsistency task their equivalents $D$ and $G$ almost always contain multiple sentences which $M$ cannot effectively handle, leading to an issue known as the granularity problem \cite{utama2022falsesum}. To bypass the granularity problem, a natural generalization is to split both $D$ and $G$ into sentences and run $M$ pairwise on each of those sentences then using an aggregation function $f$ to generate the final entailment probability. Numerous papers have used this approach to generate competitive results \cite{schuster2022stretching, laban2022summac} however this generalization is hindered by a few shortcomings.

First, the sentence decomposition of $D$ and $G$ does not properly capture the context provided in $D$. By decomposing $D$ and $G$ into single sentences $D=(d_1, d_2, \ldots,d_i,\ldots, d_{|D|})$ and $G=(g_1, g_2, \ldots,g_j,\ldots, g_{|G|})$ and put into the model to evaluate as $M(d_i, g_j)$, the context and long term dependencies present in $D$ that may have factually supported $g_j$ very likely could not be represented in $d_i$. Multiple sentences (e.g., $\cup_{i\in\{1,3,6\}}d_i$) together in unison may be needed to support a single claim $g_j$. However, evaluating $g_j$ against each sentence individually $M(d_1, g_j)$, $M(d_3, g_j)$, $M(d_6, g_j)$ would likely lead to artificially low scores. Second, evaluating pairwise sentences of $D$ and $G$ is slow. It requires $|D|\cdot |G|$ model runs to obtain a final score for one sentence $g_j$.
 
% However, this generalization has two shortcomings. First, it's not capturing context. When decomposing document $D$ into single facts or sentences $s_j$ to run against a hypothesis $h\in G$ the context and long term dependencies that may support a hypothesis $h$ are lost. If hypothesis $h$ is supported in document $D$ by $h=\cup_{i\in\{1,3,6\}}s_i$ then $\{M(s_i, h) \text{s.t.} i\in\{1,3,6\}\}$ will give poor results while $M(\cup_{i\in\{1,3,6\}}s_i, h)$ would give positive results. Second, it's slow. Model $M$ must be run $|D|\cdot |G|$ times in order to gain an answer for each fact contained in $G$. Ideally the number of  model runs it takes to evaluate a hypothesis $h\in G$ would be kept to a minimum, if not 1.

\sysname~poses a different solution to the granularity problem by decomposing $D$ into much larger chunks, which can be visualized in Figure \ref{fig:chunking}. Formally, \sysname~decomposes document $D$ into a set of $N$ chunks $C=c_1, c_2, \ldots, c_N$ such that $\cup_{c\in C}=D$. \sysname~can handle chunks of arbitrary length only limited by memory requirements, thus drastically increasing the context window provided to the model through the premise. The generated text $G$ is broken into sentences $G=(g_1, g_2, \ldots,g_j,\ldots, g_{|G|})$. We propose that decomposing $D$ using chunks rather than sentences does not negatively affect the granularity problem but rather enables superior contextual capture in model $M$, boosts accuracy, and requires significantly less model runs.

\sysname~uses Flan-T5 \cite{chung2022scaling} as a backbone NLI model $M$.  
\sysname~obtains the probability that a chunk $c_i$ entails a generated sentence $g_j$ through the following steps. First, logits are obtained by prompting $M$ with the following:
$\text{logits}=M($``\{$c_i$\} Question: does this imply `\{$g_j$\}'? Yes or no?''$)$. 
The entailment probability between $g_j$ and $c_i$ is then calculated by 
\begin{align*}
    P_{\text{entail}}=\text{SoftMax}(\text{logits}[\text{"Yes"}], \text{logits}[\text{"No"}])[0].
\end{align*}
To obtain the overall entailment score for a generated sentence $g_j$, the results are aggregated over all possible $c_i$ by,
\begin{align*}\small
    \text{SCALE}(C,g_j)=\max_{i=1 \ldots N}(P_{\text{entail}}(c_i, g_j))
\end{align*}
to obtain a final measure of factual consistency.

% Similar to previous works, we consider neutral as part of the contradictory category. Consistency can be thought of as very similar to the NLI formulation of entailment, we are essentially testing that the text is strictly a factual subset of the source document.

% Our approach as shown in Figure \ref{fig:chunking} uses chunking, a sliding window to break the source document into large portions of text we call chunks. 

% Chunking allows SCALE to stretch the premise length past simply sentences, giving SCALE more context and the ability to resolve co-reference resolutions between sentences.
% We use generated text as the NLI hypothesis and evaluate at the sentence level similar to \cite{utama2022falsesum} in order to better align with NLI pretraining and limit the amount of factual statements that must be tested for consistency.
% Each hypothesis sentence is evaluated with every premise chunk.
% \xz{If I was a reviewer I'd be curious to see how the $chunk\_size$ impacts performances in the Experiments.}

\begin{figure}[t]
    \centering
    \resizebox{\columnwidth}{!}{%
    \includegraphics{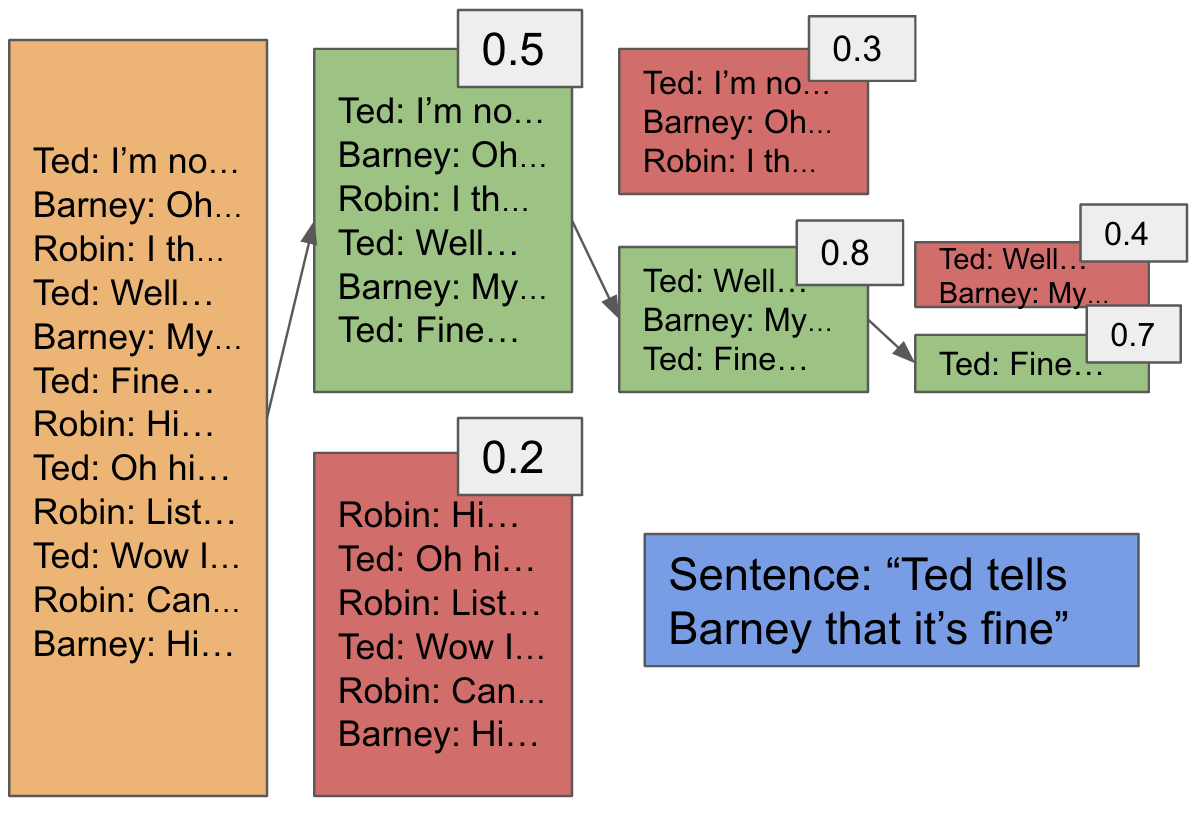}}
    \caption{Visualization of SCALE's in context retrieval using chunks to find the most relevant source utterance given a sentence. Each chunk is scored by SCALE, as shown by the gray boxes.}
    \label{fig:retrieval}
\end{figure}

\subsection{Model Explanation via Relevant Source Text Retrieval}
To produce explainable and interpretable scores for factual consistency and empower necessary human inspection, it is important to justify the score by retrieving relevant text from the source. Formally, the retrieval task involves finding the most relevant sentence $d_i$ in document $D$ with respect to a hypothesis $h$. Using a new search tree approach enabled by chunking, \sysname~is able to retrieve $d_i$ in context while using far fewer model runs than previous approaches.
We use a greedy search tree approach that evaluates a hypothesis using \sysname~against progressively smaller chunks to find the highly relevant text from the source document.
For the following example, assume we use a binary search tree (BST) at each level, dividing the text into two chunks. This process can be visualized in Figure \ref{fig:retrieval}.
Given a hypothesis $h$, we want to find the most relevant utterance $d_i$ in the source text. 
We begin by dividing the entire source document into two large chunks.
\sysname~is used to calculate the score between both chunks and the hypothesis $h$ and then use the higher scoring chunk as the new source text.
The new source text is then divided into two chunks, and continues to descend in this manner until the chunk size becomes a single sentence or utterance $d_i$.
The best scoring chunk is then chosen to be the supporting proof of the hypothesis from the source document. 

This retrieval approach is able to significantly reduce the number of model calls needed to find the relevant text from a source document.
Previous approaches commonly break the source document down by sentence which requires $\text{O}(n)$ model calls for a source document with $n$ sentences.
Whereas our BST approach only needs $\text{O}(\log(n))$ model calls in order to find the most relevant utterance in the same source document of $n$ sentences. 

Notice that we proposed the binary search scheme due to its simplicity and its connection to the popular binary search tree. In practice, diving the source text into only two chunks might cause out of GPU memory issues. In this case, we could generalize the proposed approach into different chunk splits. For example, we could divide the remaining tokens into three chunks or larger for the search of each block until the model fits the chunk. We could also use different chunk sizes for different layers so long as it fits in the memory.

\section{ScreenEval Dataset}\label{Sec:data}
We introduce a novel dataset for evaluating inconsistency detection on long form dialogues called ScreenEval.
This dataset uses TV scripts and summaries pulled from the SummScreen \cite{chen2021summscreen} dataset.
In addition to the provided human summaries, we generate summaries using Longformer and GPT-4 on 52 scripts from the SummScreen test set. We then hire human annotators to classify the factual inconsistency of each summary sentence and identify relevant supporting utterances for factually consistent summary sentences.
ScreenEval is released publicly. Details of how we use Longformer and GPT-4 and collect human annotation can be found in the Appendix \ref{appendix:dataset_construction}.

The SummScreen dataset is comprised of 2 sub datasets pulled from different sources ForeverDreaming and TVMegaSite. We use the ForeverDreaming subset of SummScreen, (SummScreen-FD) to create ScreenEval due to its manageable summary size and diversity of shows and genres, spanning a total of 21 genres.
SummScreen-FD uses human-written gold summaries from Wikipedia and TVMaze. Table \ref{tab:data_stats} shows statistics related to ScreenEval.
Notably, the average number of tokens in a source document is 6,073, which, to the best of our knowledge, makes ScreenEval the longest dialogue based inconsistency detection dataset created.
We provide 52 documents with an associated 624 summary sentences, 455 of which are artificially generated using Longformer and GPT-4.
Summaries are kept at a high level, covering major plot points and character developments.
This leads to shorter and more concise summaries that on average only run 101 tokens.

% We create summaries for ScreenEval using two summarization models, GPT-4 and Longformer, as well as human generated summaries.
% The models that we use for summarization are designed to have a large token context window, giving them both the ability to globally attend over the long source dialogues we provide.

\begin{table}[t]
    \centering
    \footnotesize
    \resizebox{\columnwidth}{!}{%
    \begin{tabular}{|c|c|}
        \hline
        Metric & Count \\
        \hline
         \# of documents & 52 \\
         \# of summary sentences & 624 \\
         avg. \# of utterances per doc & 309 \\
         avg. \# of sentences per summary & 4 \\
         \hline
         avg. \# of tokens per doc & 6073 \\
         avg. \# of tokens per summary & 101 \\
         avg. \# of tokens per summary sentence & 26 \\
         \hline
         \# factually consistent sentences & 168 \\
         \# factually inconsistent sentences & 58 \\
         avg. \# of relevant utterances & 5 \\
         \hline
    \end{tabular}}
    \caption{Statistics for the ScreenEval dataset. These statistics use the Flan-T5 tokenizer}
    \label{tab:data_stats}
\end{table}

\section{Experiments}
% \xz{Changed the section structure a bit here.}
\subsection{Datasets}
\paragraph{TRUE}
The TRUE benchmark contains 11 datasets from 4 different NLG tasks. We compare our approach to others on this dataset to show how \sysname~performs across a multitude of NLG tasks on inconsistency detection. Notably, the average number of tokens per example in the TRUE benchmark is small, generally less than 512. Each dataset in the TRUE benchmark is condensed down into a source document, generated text, and factual inconsistency label. The datasets are distributed across different tasks as shown in Table \ref{tab:task_distribution}.

\begin{table}[h]
  \centering
  \resizebox{.75\columnwidth}{!}{%
  \begin{tabular}{l | c | c}
    \textbf{Task} & \textbf{Examples} & \textbf{Datasets}\\
    \hline
    Summarization & 5,245 & 5 \\
    Dialogue & 10,613 & 3 \\
    Fact Verification & 81,263 & 2 \\ 
    Paraphrasing & 8,000 & 1 \\
  \end{tabular}}
  \caption{Number of examples and datasets for each task in the TRUE benchmark.}
\label{tab:task_distribution}
\end{table}

Summarization datasets are from FRANK \cite{pagnoni-etal-2021-understanding}, SummEval \cite{fabbri2021summeval}, MNBM \cite{maynez-etal-2020-faithfulness}, QAGS-CNNDM \cite{wang-etal-2020-asking}, and QAGS-XSum \cite{wang-etal-2020-asking}. Dialogue datasets include BEGIN \cite{dziri2022evaluating}, $Q^2$ \cite{honovich-etal-2021-q2}, and DialFact \cite{gupta2021dialfact}. Fact Verification datasets are FEVER \cite{thorne-etal-2018-fact} and VitaminC \cite{schuster-etal-2021-get}. The Paraphrasing dataset is PAWS \cite{zhang-etal-2019-paws}.

\paragraph{ScreenEval}
Our datasest ScreenEval compares the inconsistency detection ability of methods on long form dialogue. Details can be found in Sec. \ref{Sec:data}

\subsection{Competitive Systems}
TRUE provides 9 inconsistency detection baselines from 4 different inconsistency detection styles, namely n-gram based methods (token level F1), model based methods (BERTScore \cite{zhang2019bertscore}, BLEURT \cite{sellam-etal-2020-bleurt}, FactCC \cite{kryscinski-etal-2020-evaluating}, BARTScore \cite{yuan2021bartscore}, CTC \cite{deng-etal-2021-compression}), NLI based methods (ANLI \cite{honovich2022true}, SummaC \cite{laban2022summac}), and question answering (QA) based methods ($\text{Q}^2$ \cite{honovich-etal-2021-q2}, QuestEval \cite{scialom-etal-2021-questeval}).

For ScreenEval we compare \sysname~to 8 models which use NLI, QA, modern GPT systems, and older n-gram and semantic similarity methods. The baseline models consist of two NLI based sentence decomposition approaches seNtLI \cite{schuster2022stretching}, and $\text{SummaC}_{conv}$ \cite{laban2022summac}, a state-of-the-art QA based model QAFactEval \cite{fabbri-etal-2022-qafacteval}, a multidimensional QA model UniEval \cite{zhong2022towards}, a semantic similarity model based method BERTScore \cite{zhang2019bertscore}, an n-gram overlap method ROUGE \cite{lin2004rouge}, and the two recent OpenAI models ChatGPT, and GPT-4. 
% Similar to our approach, SummaC is a NLI based metric.
% SummaC and seNtLI use a sentence by sentence comparison approach to handle arbitrarily large documents.
% We compare SCALE to the two versions of SummaC (Zero Shot and Conv) as well as seNtLI in order to directly contrast chunking and sentence decomposition NLI based approaches. 
% UniEval is a multidimensional QA model that is in part fine tuned for consistency detection.
% UniEval uses T5 which is similar to our backbone model Flan-T5.
% BERTScore is a semantic similarity metric that uses BERT to compare the embeddings of two given texts.
% Rouge is a n-gram based metric that is commonly used to evaluate consistency detection, and is thus useful baselines to compare against.
% ChatGPT and GPT-4 are LLMs that have powerful zero shot capabilities also using natural language prompts. We prompt ChatGPT and GPT-4 with an NLI inspired yes/no question to have it classify the factual inconsistency of summary sentences in ScreenEval. One drawback of GPT-4 and ChatGPT is that their outputs are discrete rather than continuous, meaning they cannot be measured for calibration and subtle improvements may not be captured in their binary labelling.

We also compare the performance of SCALE's search tree based relevant utterance retrieval with other recent retrieval models.
We compare \sysname~to the retrieval performance of SuperPal \cite{ernst2020summary} which was shown to have superior retrieval capabilities in LongEval.
We also compare against seNtLI \cite{schuster2022stretching} which was designed to perform retrieval to identify factual inconsistencies over long documents.

For \sysname~, we include three variants of Flan-T5 as the backbone, namely base, XL, and XXL.

\begin{table*}[ht!]
\centering
\footnotesize
\resizebox{\textwidth}{!}{%
\begin{tabular}{|*{13}{>{\centering\arraybackslash}l|}}
\hline
Metric & FRANK & SummEval & MNBM & QAGS-C & QAGS-X & BEGIN & Q\textsuperscript{2}\textsubscript{ds} & DialFact & PAWS & FEVER & VitC & Avg\textsubscript{\text{w/o VitC,FEVER}} \\
\hline
Q\textsuperscript{2}\textsubscript{metric} & 87.8 & 78.8 & 68.7 & 83.5 & 70.9 & 79.7 & 80.9 & 86.1 & 89.7 & 88.4 & 81.4 & 80.7 \\
\hline
ANLI & 89.4 & 80.5 & $\textbf{77.9}$ & 82.1 & 83.8 & 82.6 & 72.7 & 77.7 & 86.4 & 93.2 & 88.3 & 81.5 \\
\hline
SC\textsubscript{ZS} & 89.1 & 81.7 & 71.3 & 80.9 & 78.1 & 82.0 & 77.4 & 84.1 & 88.2 & \sout{93.2} & \sout{97.9} & 81.4 \\
\hline
F1 & 76.1 & 61.4 & 46.2 & 63.8 & 51.1 & 86.4 & 65.9 & 72.3 & 51.1 & 51.8 & 61.4 & 63.8 \\
\hline
BLEURT & 82.8 & 66.7 & 64.5 & 71.6 & 57.2 & 86.4 & 72.4 & 73.1 & 68.3 & 59.5 & 61.8 & 71.4 \\
\hline
QuestEval & 84.0 & 70.1 & 65.3 & 64.2 & 56.3 & 84.1 & 72.2 & 77.3 & 69.2 & 72.6 & 66.5 & 71.4 \\
\hline
FactCC & 76.4 & 75.9 & 59.4 & 76.4 & 64.9 & 64.4 & 63.7 & 55.3 & 64.0 & 61.9 & 56.3 & 66.7 \\
\hline
BART\textsubscript{score} & 86.1 & 73.5 & 60.9 & 80.9 & 53.8 & 86.3 & 64.9 & 65.6 & 77.5 & 64.1 & 63.2 & 72.2 \\
\hline
BERT\textsubscript{score} & 84.3 & 77.2 & 62.8 & 69.1 & 49.5 & 87.9 & 70.0 & 64.2 & 77.5 & 63.3 & 62.5 & 71.4 \\
\hline \hline
SCALE\textsubscript{large} & 88.0 & 79.4 & 74.2 & 81.0 & 81.4 & 79.3 & 84.3 & 91.6 & 96.9 & 93.9 & 89.3 & 84.0 \\
\hline
SCALE\textsubscript{xl} & 89.8 & 83.3 & 73.2 & 82.1 & 84.3 & 77.8 & 86.3 & 91.0 & 91.9 & 94.5 & 91.5 & 84.4 \\
\hline
SCALE\textsubscript{xxl} & \textbf{90.8} & \textbf{91.5} & 73.1 & \textbf{85.2} & \textbf{85.3} & \textbf{79.2} & \textbf{86.6} & \textbf{93.4} & \textbf{96.7} & \textbf{94.8} & \textbf{92.7} & \textbf{88.1} \\
\hline
\end{tabular}}
\caption{TRUE benchmark results. ROC\_AUC scores multiplied by 100 for readability. Since $SC_{ZS}$ uses FEVER and VitC in training, these two datasets are excluded when computing the average.}
\label{tab:true}
\end{table*}

\subsection{Metrics}

\paragraph{Accuracy Evaluation} 
We compare the performance of methods primarily using four metrics, ROC\_AUC score, Pearson correlation, Kendall\_Tau correlation, and F1\_Macro score. We employ the ROC\_AUC score to quantify the ability of different methods in accurately identifying true positives and true negatives. Pearson and Kendall\_Tau correlations show the relationship between methods and labels by measuring the correlations between the two. Finally the F1\_Macro score is used to compare the continuous outputs of \sysname~to the discrete outputs of GPT-4 and ChatGPT. To obtain an F1 score for \sysname~we use the optimal threshold to convert its continuous output into discrete values.

\paragraph{Efficiency Evaluation}
We measure wall clock time in seconds for all of our experiments on ScreenEval. Wall clock time demonstrates how \sysname~can be used efficiently in an online setting especially when compared to other models. 

\paragraph{Model Explanation Evaluation}
We evaluate Calibration and Relevant Source Text Retrieval for model explanation. 

% \xz{This subsection is an introduction of ECE, it should be under Experiments instead of our methods.} \pat{Second this. And it's also not very intuitive why we need to include calibration here. And even if we have a better calibration. People want to know why.}
\textbf{Calibration} is the measure of how close the pseudo-probability outputs of a model are to the actual probability of a correct prediction. 
For example, a well calibrated model that produces a score of $0.2$ would have a $20\%$ probability of being classified as 1.
 In this paper, we use Expected Calibration Error (ECE) \cite{guo2017calibration} to compare the calibration of \sysname~to other commonly used models.

Given model outputs spanning from 0 to 1, ECE separates the outputs into $K$ 
equally sized bins $B_k$ between 0 and 1 and takes the difference between accuracy $acc$ and confidence $conf$ in each one.
The accuracy of a bin $B_k$ is the average amount of predicted labels that match true class labels in a bin, formally defined as
\begin{equation}
    acc(B_k) = \frac{1}{|B_k|} \sum_{i\in B_k} \textbf{1}(\hat{y_i}=y_i),
    \label{eq:acc}
\end{equation}
where $\hat{y_i}$ and $y_i$ are the predicted and true class labels for sample $i$. Confidence in a bin $B_k$ shows the average predicted score in a bin, formally defined as
\begin{equation}
    conf(B_k) = \frac{1}{|B_k|} \sum_{i\in B_k} \hat{p_i},
    \label{eq:conf}
\end{equation}
where $\hat{p_i}$ is the model output score for sample $i$.

Then the following equation is used to calculate ECE,
\begin{equation}
    ECE=\sum^K_{k=1}\frac{|B_k|}{n}|acc(B_k)-conf(B_k)|,
\end{equation} 
using equation (\ref{eq:acc}) and (\ref{eq:conf}). A lower ECE indicates a better calibration.

\textbf{Relevant Source Text Retrieval} tests if each model could return the correct utterance identified as relevant by human labelers. We report the recall of retrieval results.

\begin{table*}[ht!]
\centering
\footnotesize
\resizebox{\textwidth}{!}{%
\begin{tabular}{|*{13}{>{\centering\arraybackslash}l|}}
\hline
Metric & FRANK & SummEval & MNBM & QAGS-C & QAGS-X & BEGIN & Q\textsuperscript{2}\textsubscript{ds} & DialFact & PAWS & FEVER & VitC & Avg \\
\hline
UniEval & 0.333 & 0.238 & 0.173 & 0.26 & 0.315 & 0.507 & 0.153 & 0.396 & 0.335 & 0.084 & 0.242 & 0.276 \\
\hline
BERTScore & 0.2343 & 0.0730 & 0.1568 & 0.3200 & 0.4783 & 0.3119 & 0.1760 & 0.1388 & 0.1923 & 0.4217 & 0.0648 & 0.233 \\
\hline
SummaC\textsubscript{c} & 0.199 & 0.322 & 0.128 & 0.067 & \textbf{0.043} & 0.085 & 0.412 & 0.116 & 0.355 & 0.109 & 0.185 & 0.184 \\
\hline
SummaC\textsubscript{z} & 0.266 & 0.082 & 0.058 & 0.267 & 0.196 & 0.244 & \textbf{0.084} & 0.08 & 0.122 & \textbf{0.04} & 0.074 & 0.137 \\
\hline
SCALE\textsubscript{xl} & \textbf{0.093} & 0.243 & \textbf{0.026} & \textbf{0.060} & 0.116 & \textbf{0.069} & 0.286 & \textbf{0.071} & 0.060 & 0.209 & 0.150 & 0.126 \\ 
\hline
SCALE\textsubscript{large} & 0.179 & \textbf{0.043} & 0.215 & 0.144 & 0.117 & 0.149 & 0.103 & 0.117 & \textbf{0.019} & 0.058 & \textbf{0.049} & \textbf{0.108} \\ 
\hline
\end{tabular}}
\caption{TRUE Calibration Results. ECE of each method on TRUE benchmark datasets (lower is better)}
\label{tab:true_calibration}
\end{table*}

\section{Results}
% \xz{Do we have human evaluation results to show our metric correlates the best in terms of consistency? And an analysis on how the $chunk_size$ impacts the performances?}

\subsection{TRUE}
We first evaluate \sysname~on the TRUE benchmark to confirm \sysname~is NLG task agnostic and generalizes well to the factual inconsistency detection. 

\paragraph{Accuracy Evaluation Results}
For the TRUE benchmark as shown in Table \ref{tab:true}, $\text{SCALE}_{XXL}$ provides superior performance in 10 out of the 11 datasets, and $\text{SCALE}_{XL}$ achieves superior performance in 8 datasets compared to other non-\sysname~models. 
Notably, other models were not previously able to perform well across all tasks, with $Q^2_{metric}$ having superior performance across 3 datasets and ANLI having superior performance across 5. 
These results demonstrate SCALE's ability to perform well across domains and against a large variety of model types.

% Through \sysname~achieving this level of performance without decomposing the premise we support our claim that the granularity problem only applies to the hypothesis and not the premise in an NLI setting.
% We find instead that a lack of decomposing the premise leads to better results and requires less runs of the model.

\paragraph{Model Explanation Evaluation Results}

Not only does \sysname~provide superior performance on the TRUE benchmark, but it is also highly calibrated across NLG tasks. 
Table \ref{tab:true_calibration} shows the ECE of multiple methods across the TRUE benchmark datasets. 
Note that a lower ECE is better. 
$\text{SCALE}_{large}$ provides the best calibration on average and $\text{SCALE}_{XL}$ outperforms other non-\sysname~models in calibration on over half of the datasets in the TRUE benchmark. 

\begin{figure}[t]
    \centering
    \resizebox{\columnwidth}{!}{%
    \includegraphics{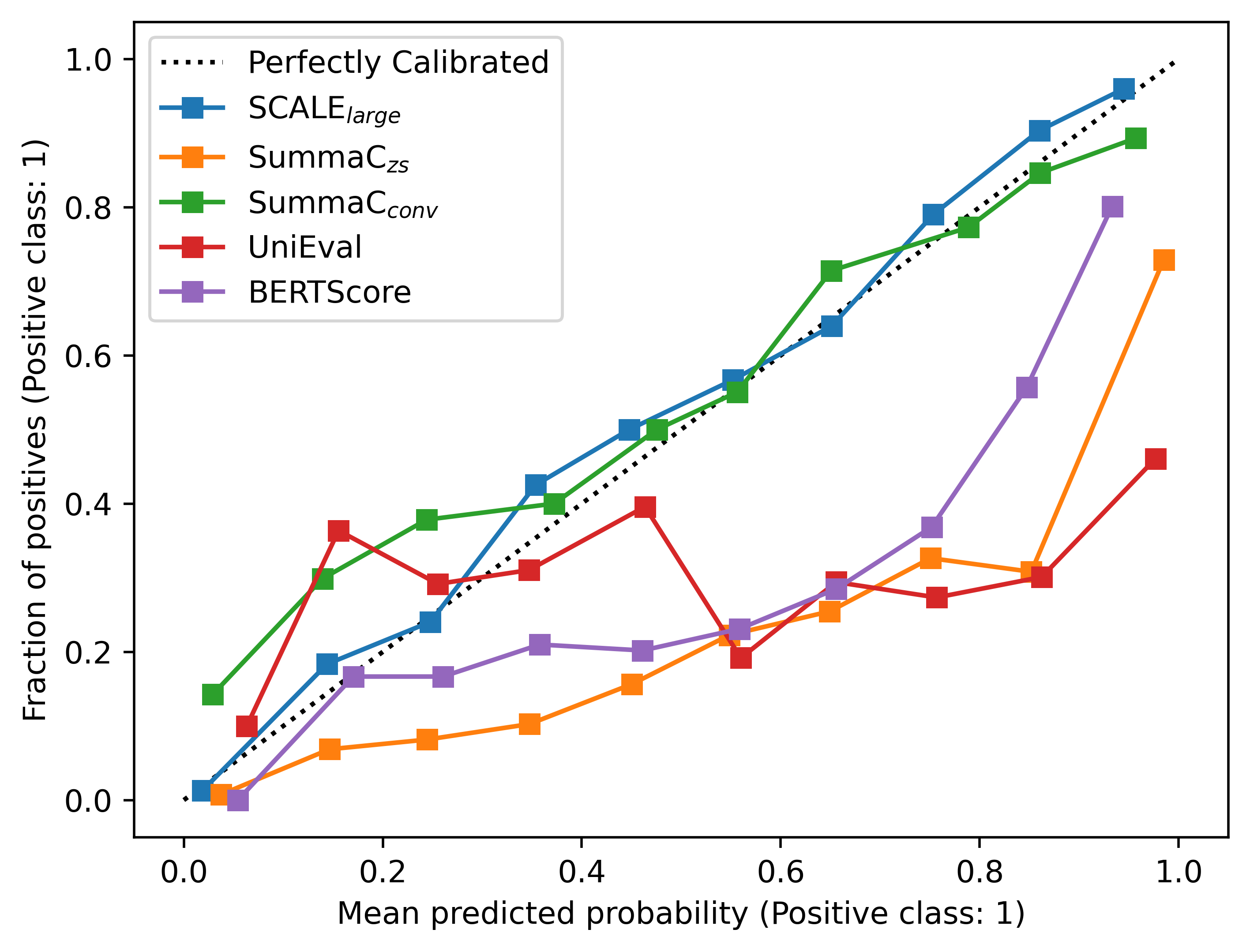}}
    \caption{Calibration curves on the PAWS benchmark}
    \label{fig:cal_curve_true}
\end{figure}

\begin{figure}[h]
    \centering
    \resizebox{\columnwidth}{!}{
    \includegraphics{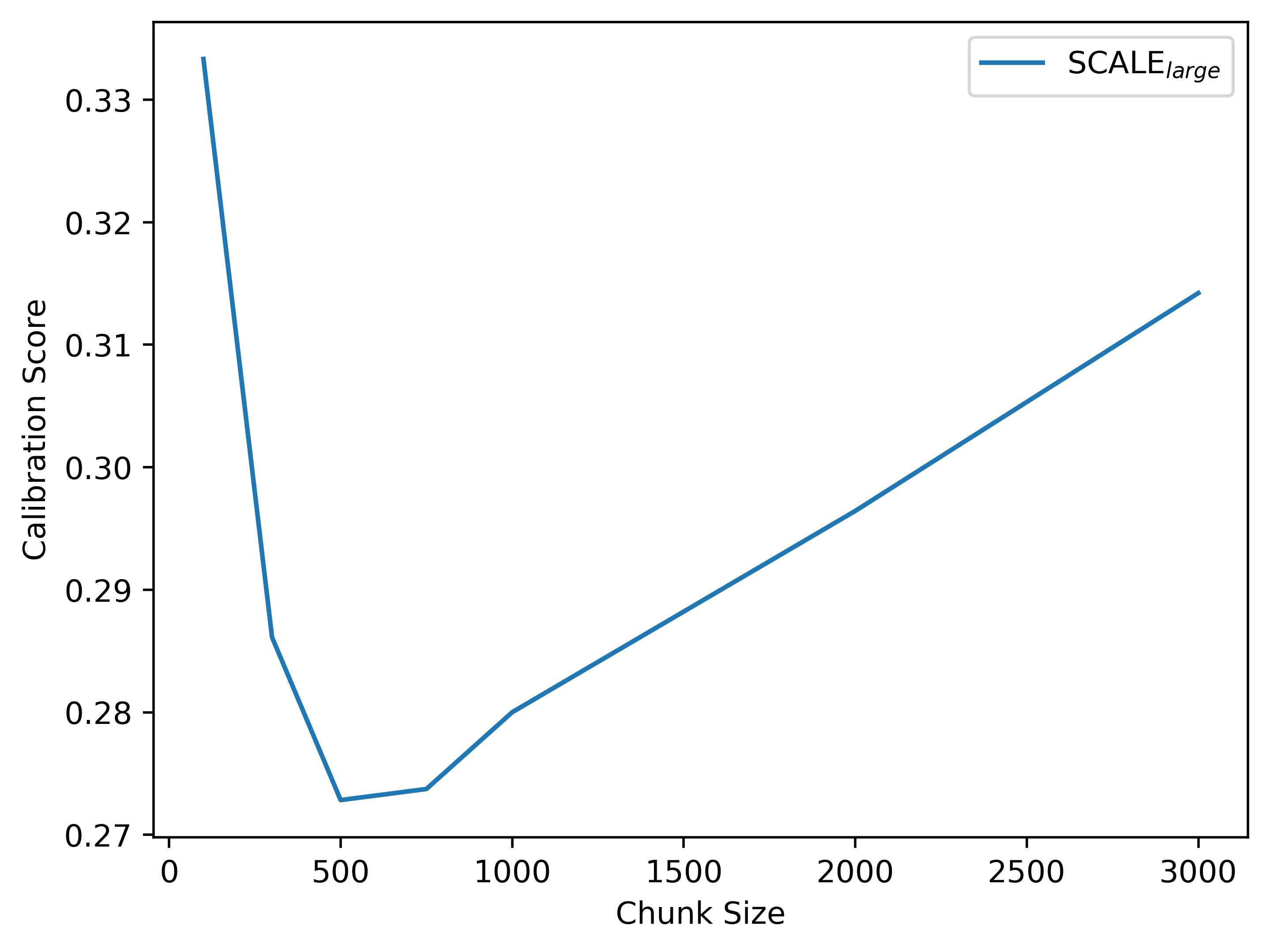}}
    \caption{Effect of different chunk sizes on calibration performance on ScreenEval dataset.}
    \label{fig:calibration_exp}
\end{figure}

A visual example of the calibration results can be analyzed with the calibration curves in Figure \ref{fig:cal_curve_true}. 
While most models are severely uncalibrated and underestimate the fraction of positives in Figure \ref{fig:cal_curve_true}, \sysname~is capable of sticking extremely close to the perfectly calibrated line. The closest model $\text{SummaC}_{Conv}$ can be seen overestimating positive examples before scores reach $0.4$. We hypothesize that the large context window is the key to better calibration in SCALE as it includes more information. This makes the underlying NLI model less likely to be biased toward a specific range of tokens which leads to extreme confidence based on certain short text. To empirically justify this, we perform further experiments on the proposed ScreenEval dataset shown in Figure \ref{fig:calibration_exp}. We can observe that for chunk size $<$ 400, the calibration score (the lower the better) is much higher than larger chunk size 500 to 1000. This shows that a larger chunk size could enable the NLI model to extract more useful information to provide appropriate confidence when making the prediction. We also use this knowledge to support our decision to use 512 tokens as our chunk size for all experiments in this paper.
The enhanced calibration achieved by \sysname~allows it to be more interpretable as a probability, making it a valuable tool for comparison tasks.

\subsection{ScreenEval}
We then evaluate the performance of SCALE's chunking capabilities against other models in a long form dialogue setting using ScreenEval. We evaluate the factual inconsistency detection and relevant utterance retrieval of \sysname~compared with other models and explore the unique problems posed by long form dialogue evaluation.

\begin{table}[ht!]
\centering
\footnotesize
\resizebox{\columnwidth}{!}{%
\begin{tabular}{|*{5}{>{\centering\arraybackslash}l|}}
\hline
Metric & Pearson & Kendall-Tau &  ROC-AUC & Time (s) \\
\hline
Rouge-1 & 0.120 & 0.093 &  57.6 & - \\
\hline
BERTScore & 0.130 & 0.076 &  56.2 & - \\
\hline
SummaC\textsubscript{c} & 0.030 & 0.011 &  50.9 & 1153\\
\hline
UniEval & 0.066 & 0.094 &  57.6 & 1139\\
\hline
seNtLI & 0.005 & 0.096 &  42.2 & 12688\\
\hline
QAFactEval & 0.331 & 0.293 & 73.5 & 12132\\ 
\hline
SCALE\textsubscript{base} & 0.28 & 0.24 & 69.5 & \textbf{678}\\ 
\hline
SCALE\textsubscript{large} & \textbf{0.391} & \textbf{0.322} & \textbf{76.1} & 1991\\ 
\hline
\end{tabular}}
\caption{Factual inconsistency detection results on ScreenEval. ROC\_AUC is multiplied by 100 for readability.}
\label{tab:our_con}
\end{table}

\paragraph{Accuracy and Efficiency Evaluation Results}
We compare the factual inconsistency detection performance of multiple models on ScreenEval in Table \ref{tab:our_con}. \sysname~significantly outperforms other methods across all measures. While the state-of-the-art QA model QAFactEval was able to perform well on ScreenEval, $\text{SCALE}_{large}$ still showed superior performance across all metrics. Notably, even $\text{SummaC}_{conv}$ and seNtLI, which are designed to deal with long documents, have poor performance on ScreenEval.

Along with its superior performance, \sysname~is able to run faster than other LLM based methods on ScreenEval also shown in Table \ref{tab:our_con}. For a fair comparison, we set the batch size to 1 for all models and run with all other default settings. We do not include BERTScore due to it's truncation of the document, making timing not comparable. 
Most notably QAFactEval, which was closest in performance to $\text{SCALE}_{large}$, was 6 times slower than $\text{SCALE}_{large}$ in wall clock time. Even faster though was $\text{SCALE}_{base}$ which was 17 times faster than QAFactEval while only achieving slightly worse performance across all metrics on ScreenEval, and outperforming all non-SCALE methods other than QAFactEval. The $\text{SCALE}_{base}$ model running at 1.1 seconds per score for long documents could realistically be used in an online setting to more accurately evaluate factual inconsistency. 

\begin{figure}[h]
    \centering
    \resizebox{\columnwidth}{!}{
    \includegraphics{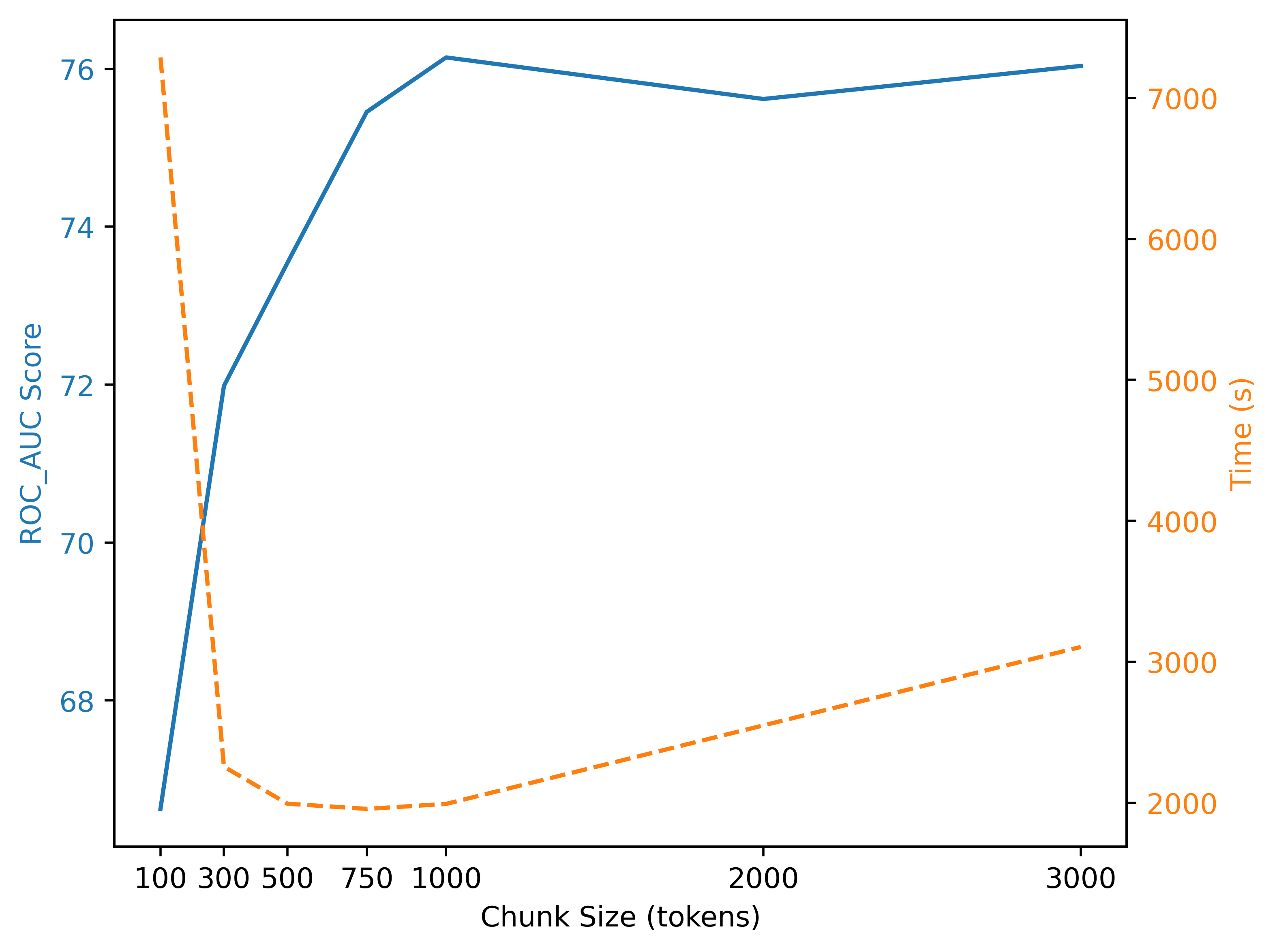}}
    \caption{Effect of different chunk sizes on $\text{SCALE}_{large}$ performance and time on ScreenEval dataset. ROC$\_$AUC score is multiplied by 100 for readability.}
    \label{fig:chunk_size_and_time}
\end{figure}

Chunk size proves to have a large effect on the ability of SCALE's performance and time as seen in Figure \ref{fig:chunk_size_and_time}. $\text{SCALE}_{large}$ sees a sharp increase in performance up until the chunk size is 1000 tokens long. Similarly, there is a sharp decrease in model run time up until 1000 tokens. Figure \ref{fig:chunk_size_and_time} substantiates our approach to the granularity problem by illustrating that a larger number of tokens in the premise leads to a more effective method.

\begin{table}[h]
\centering
\resizebox{\columnwidth}{!}{%
\begin{tabular}{|l|c|c|c|}
\hline
Model & F1-score macro & Cost & Time (s) \\ \hline
SCALE\textsubscript{XL}      & 73.86                        & -             & 5425             \\ \hline
SCALE\textsubscript{large}   & 68.74                        & -             & \textbf{1991}              \\ \hline
GPT-4           & \textbf{77.95}                        & \$102         & 7255             \\ \hline
ChatGPT        & 56.5                         & \$5           & 2544              \\ \hline
\end{tabular}}
\caption{Compare \sysname~with ChatGPT and GPT-4 on ScreenEval. }
\label{table:gpt_compare}
\end{table}

We additionally compare \sysname~with ChatGPT and GPT-4 on ScreenEval in Table \ref{table:gpt_compare}. Due to the discrete nature of GPT-4 and ChatGPT's output, we choose an ideal threshold for \sysname~and compare macro F1-Scores on the ScreenEval dataset. While GPT-4 is able to outperform $\text{SCALE}_{XL}$ in macro F1-Score, \sysname~shows to be significantly better in terms of time and cost. ChatGPT is more comparable in terms of time and cost to SCALE; however, there is a significant performance drop in macro F1-Score. ChatGPT is also limited by its 4096 token length limit at the time of writing and must use truncated conversations from ScreenEval. These results help us conclude that while GPT-4 has superior performance, \sysname~is able to provide a faster, more affordable model that can be used locally in an online setting to produce continuous scores. 
\paragraph{Model Explanation Evaluation Results}
\begin{table}[ht!]
\centering
\footnotesize
\resizebox{.75\columnwidth}{!}{
\begin{tabular}{|*{3}{>{\centering\arraybackslash}l|}}
\hline
Metric & Recall & Time (s) \\
\hline
seNtLI\textsubscript{xl} & 0.343 & 35.35 \\
\hline
SuperPal & 0.406 &  6.13 \\ 
\hline
SCALE\textsubscript{large} & 0.463 &  \textbf{4.67} \\ 
\hline
SCALE\textsubscript{xl} & \textbf{0.471} &  12.46 \\ 
\hline
\end{tabular}
}
\caption{Relevant source text retrieval results on ScreenEval. Time is measured in average number of seconds taken to retrieve the most relevant utterance on ScreenEval conversations.}
\label{tab:retrieval}
\end{table}
We now compare SCALE's BST retrieval approach with SuperPal and seNtLI. SCALE outperforms both in terms of time and performance as shown in Table \ref{tab:retrieval}. 
$\text{SCALE}_{XL}$ identifies the most relevant utterances correctly 47\% of the time compared to 34\% for seNtLI and 41\% for SuperPal. 
SCALE's BST retrieval requires significantly fewer model calls, allowing it to pinpoint relevant utterances without having to score each one individually like the other methods. 
This results in higher retrieval recall for both $\text{SCALE}_{large}$ and $\text{SCALE}_{XL}$.
Moreover, because $\text{SCALE}_{large}$ requires far less model calls it is able to provide faster results than SuperPal or seNtLI without comprimising effectiveness. This enhanced performance shows how SCALE could be used in an online setting for fast and accurate results.

% \section{Future Works}
% While \sysname~performs well on long form dialogue there are numerous other avenues to explore.
% Expanding the amount of models that implement chunking and seeing how well they perform could be an interesting future avenue of exploration.
% Using simply {“Yes”, “No”} logits can lead to loss of accuracy due to other information possibly flowing to similar tokens such as {“yes”, “no”} or a pairing of different tokens such as {“yes”, “No”}. An avenue of exploration could involve exploring how one might combine these logits and if there is any benefit to doing so.
% Using NLI for multiple tasks similar to UniEval could potentially lead to a flexible multidimensional model that works well on long datasets. This paper chose to focus on factual inconsistency however it has been shown that NLG metrics can evaluate many other dimensions. It would be benificial to explore evaluating along those dimensions for long documents. 
% Long document metric evaluation is very limited to only a few datasets. It would be itnerestig to see more studies that focus on real world long document evaluations to create more effective models and metrics. 
% Calibration in metrics proves to be a difficult task and is under evaluated on new metrics. Future efforts could aim for low ECE across all NLG tasks for consistency detection and other tasks. 

\section{Conclusion}
In this paper, we introduce a cutting-edge NLI based factual inconsistency detection method called SCALE. We show that \sysname~is an NLG task agnostic factual inconsistency detection method by achieving state-of-the-art results across four distinct NLG tasks and 11 datasets within the TRUE benchmark. We show that across NLG tasks \sysname~is also superior in calibration, providing scores that are interpretable and enables accurate comparison of scores. We introduce a new dataset called ScreenEval that is the first of its kind for long form dialogue factual inconsistency evaluation. \sysname~is shown to significantly outperform all models other than GPT-4 in factual inconsistency detection on this dataset. However, we show that \sysname~is significantly more cost effective and faster than GPT-4 for evaluation on ScreenEval. Moreover, we introduce a new retrieval strategy enabled by SCALE that significantly decreases the time to retrieve relevant utterances from a long document with increased recall accuracy. 

\section{Limitations}
While SCALE performs well at inconsistency detection there are some limitations to this approach. 

SCALE only uses the "Yes" and "No" logits to compute it's entailment score, however only using those two logits specifically could lead to loss of accuracy due to other information possibly flowing to similar tokens such as {“yes”, “no”}. Using logits for scoring purposes may cause a loss of information to other similar logits.

Finally, even though SCALE is able to achieve better calibration in the aggregate, it still struggles with calibration on certain tasks and this can even vary by model size. Consistent calibration of scoring methods across NLG tasks should be a goal for future methods.
% I think limitation should be here. Instructions form EMNLP:

% Mandatory Discussion of Limitations
% We believe that it is also important to discuss the limitations of your work, in addition to its strengths. EMNLP 2023 requires all papers to have a clear discussion of limitations, in a dedicated section titled “Limitations”. This section will appear at the end of the paper, after the discussion/conclusions section and before the references, and will not count towards the page limit. Papers without a limitation section will be automatically rejected without review.

\bibliography{emnlp2023}
\bibliographystyle{acl_natbib}

\clearpage

\appendix
\section*{Appendix}

\section{Details of Construction of ScreenEval Dataset}
\label{appendix:dataset_construction}
We create summaries for ScreenEval using two summarization models, GPT-4 and Longformer, as well as human generated summaries. The models that we use for summarization are designed to have a large token context window, giving them both the ability to globally attend over the long source dialogues we provide. Below, we will first explain how to leverage Longformer, GPT-4 to generate summaries, and then explain the process and the cost of human labeling.  

\subsection{Building the Dataset}
We first generated longformer summaries for each script in the test set.
To keep the annotation task reasonable, and to filter out any rambling summaries, we filter out any TV scripts with a longformer or human summary that had more than 6 sentences or only 1 sentence.
We still preserve 52 of the TV scripts by doing this, as the median number of summary sentences in a longformer summary was 4 and a human summary was 3.
In order to meet GPT-4’s token limit requirements, from the remaining TV scripts we chose all that had less than 8,100 tokens.
Our final dataset consists of 52 TV scripts that have an average length of 6073 tokens with 624 summary sentences.

\subsection{Longformer}
We use the same baseline model as in SummScreen to generate summaries in ScreenEval, a Longformer model finetuned on SummScreen-FD's training set.
This model uses a transformer based sequence to sequence architecture to globally attend over the entire dialogue.
Longformer is able to take as many as 16384 tokens as input.

\subsection{GPT-4}
GPT-4 is a large language model that has shown near human level performance in a wide variety of tasks, including summarization.
We task GPT-4 to summarize each document in ScreenEval using the prompt
``Summarize in 5 sentences or less: ''.
To accommodate for the roughly 8k token limit on the GPT-4 model, we specifically select documents in ScreenEval that are under 8k tokens.

\subsection{Human Annotation}
\begin{figure*}[h]
    \centering
    \resizebox{\textwidth}{!}{
    \includegraphics{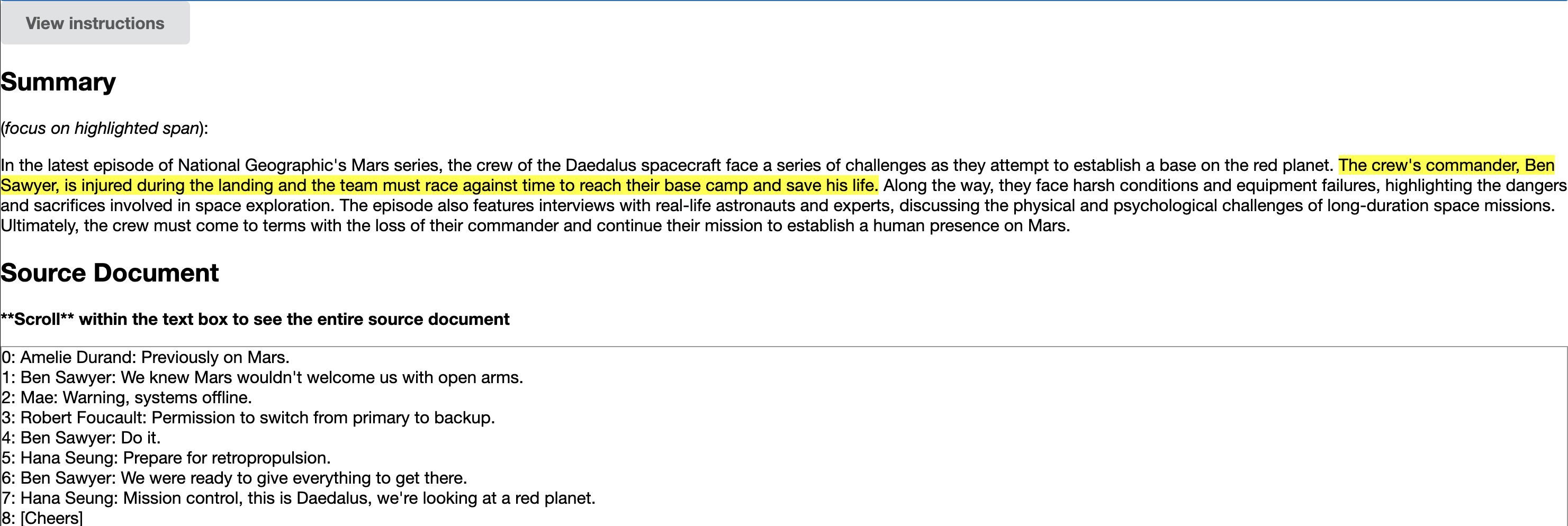}}
    \caption{Prompt provided to mturk labellers 0}
    \label{fig:label_0}
\end{figure*}

\begin{figure*}[h]
    \centering
    \resizebox{\textwidth}{!}{
    \includegraphics{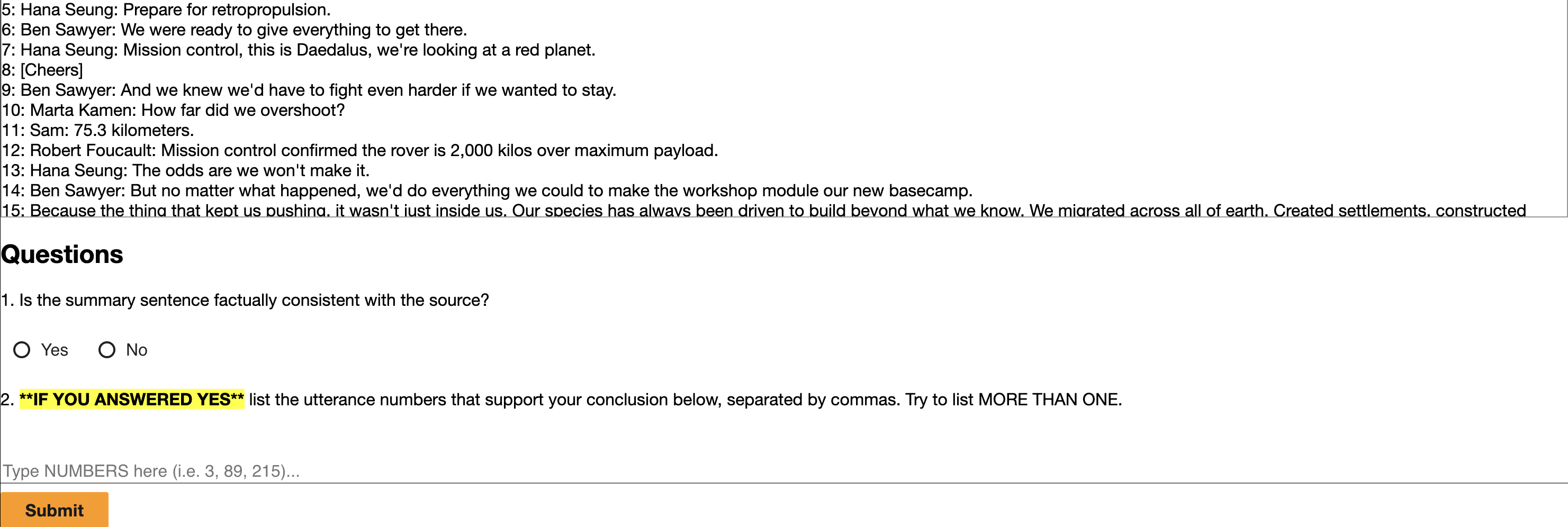}}
    \caption{Prompt provided to mturk labellers 1}
    \label{fig:label_1}
\end{figure*}

We lable ScreenEval using workers from amazon mechanical turks with the promtp shown in Figure \ref{fig:label_0} and Figure \ref{fig:label_1}.
For each task, a worker is presented with a TV script from ScreenEval along with a highlighted summary sentence.
The worker is instructed to first read through the source dialogue.
Then, the worker is instructed to click either a Yes or No button to indicate whether the highlighted summary sentence is consistent with the source document.
Each utterance in the TV script will be presented alongside a check box where if the worker chose “Yes” to the consistency question, they will be asked to select the relevant utterances that led to their answer.

We had 3 human annotators label each instance, and 61$\%$ of the time all three annotators agreed. Workers were paid 0.27 per task. We ensured the quality of annotators through a number of methods. First, we filtered annotators to just those located in the US and Canada to increase the chances of high fluency in English on our reading comprehension task. Additionally the workers had to have an MTurk “Master” qualification, greater than a 95$\%$ task approval rate, and greater than 5000 tasks approved. The dataset was labeled in batches of 30 at a time and closely monitored by the authors. Workers were only rejected if they did not list relevant utterances as instructed or listed non existent utterances, and these workers were able to dispute rejection via email.

\section{Prompts Used for GPT}
\label{sec:appendix_Prompt}
\subsection{ChatGPT/GPT-4}

\vspace{1em}
\fbox{\begin{minipage}{17em}
$\{$Dialogue$\}$

Question: does the previous conversation factually imply "$\{$Summary Sentence$\}$"? Answer Yes or No.
\end{minipage}}
\vspace{1em}

% \subsection{Cost of Labeling}
% Price estimates for the amazon mechanical turk human labeling task are as follows. The Longformer baseline has 169 summary sentences. The human baseline has 196 summary sentences. GPT-4 can summarize in ~5 sentences if provided in the prompt, meaning it will provide about 260 summary sentences. Previous works list lableling one summary as \$0.75, and we will get 3 workers to label each summary sentence. Given a 5\% additional fee for using “Master” workers and a 20\% commission fee for amazon, there is roughly a \$562.5 estimated cost if we can get labels for the \$0.75/summary price point. Another option could include paying workers \$1.50 to label all summary sentences in all three summaries, meaning they would only have to read the long source document once. This could incur an additional 20\% amazon fee for tasks that contain more than 10 assignments. We will first release 5 tasks under the first plan and evaluate how each performs. I have self labeled 8 documents with their Longformer and GPT-4 summaries on my own, each task took ~25 mins in total where 20 mins on average was dedicated to reading the TV script carefully and 5 mins to evaluate the consistency of each sentence. 

\end{document}